\documentclass{article}

    \PassOptionsToPackage{numbers, compress}{natbib}

\usepackage[main, final]{neurips_2026}

\usepackage[utf8]{inputenc} 
\usepackage[T1]{fontenc}    
\usepackage{hyperref}       
\usepackage{url}            
\usepackage{booktabs}       
\usepackage{amsfonts}       
\usepackage{nicefrac}       
\usepackage{microtype}      
\usepackage{xcolor}         

\usepackage{float}

\usepackage{caption}
\usepackage{subcaption}
\usepackage{wrapfig}
\usepackage{graphicx}

\usepackage{multirow}
\usepackage{booktabs}
\usepackage{colortbl}
\usepackage{longtable} 
\usepackage{array}

\usepackage{amsmath}
\usepackage{amssymb}
\usepackage{mathtools}
\usepackage{amsthm}
\usepackage{nicefrac} 
\usepackage{derivative} 
\usepackage{empheq} 

\usepackage{bm} 
\usepackage{mathrsfs} 
\usepackage{dsfont} 
\usepackage{pifont}
\usepackage{fontawesome5} 
\usepackage{latexsym} 
\usepackage{marvosym}  

\usepackage{lipsum}

\usepackage{comment}

\usepackage{pythonhighlight}
\usepackage{listings}

\usepackage{cleveref}

\usepackage[english]{babel}
\usepackage{csquotes}

\usepackage{enumitem}

\usepackage{url}

\usepackage{algorithm}
\usepackage{algorithmicx}
\usepackage[noend]{algpseudocode}
\makeatletter
\renewcommand{\ALG@beginalgorithmic}{\normalsize} 
\makeatother

\usepackage{tablefootnote}

\usepackage[normalem]{ulem}

\usepackage{xargs}
\usepackage[textsize=tiny]{todonotes}

\usepackage{actuarialsymbol}

\def\restrict#1{\raise-.5ex\hbox{\ensuremath|}_{#1}}

\theoremstyle{plain}
\newtheorem{theorem}{Theorem}[section]

\theoremstyle{definition}
\newtheorem{definition}[theorem]{Definition}

\theoremstyle{remark}

\newcommand{\eg}{{e.g.}}

\newcommand{\ie}{{i.e.}}

\newcommand{\one}{\mathbf{1}}  

\DeclareMathSymbol{\Alpha}{\mathalpha}{operators}{"41}
\DeclareMathSymbol{\Beta}{\mathalpha}{operators}{"42}
\DeclareMathSymbol{\Epsilon}{\mathalpha}{operators}{"45}
\DeclareMathSymbol{\Zeta}{\mathalpha}{operators}{"5A}
\DeclareMathSymbol{\Eta}{\mathalpha}{operators}{"48}
\DeclareMathSymbol{\Iota}{\mathalpha}{operators}{"49}
\DeclareMathSymbol{\Kappa}{\mathalpha}{operators}{"4B}
\DeclareMathSymbol{\Mu}{\mathalpha}{operators}{"4D}
\DeclareMathSymbol{\Nu}{\mathalpha}{operators}{"4E}
\DeclareMathSymbol{\Omicron}{\mathalpha}{operators}{"4F}
\DeclareMathSymbol{\Rho}{\mathalpha}{operators}{"50}
\DeclareMathSymbol{\Tau}{\mathalpha}{operators}{"54}
\DeclareMathSymbol{\Chi}{\mathalpha}{operators}{"58}
\DeclareMathSymbol{\omicron}{\mathord}{letters}{"6F}

\newcommandx{\yihan}[2][1=]{\todo[linecolor=teal,backgroundcolor=teal!25,bordercolor=teal,#1]{#2}}

\newcommand{\Ebb}{\mathbb{E}}

\newcommand{\mbf}{\mathbf{m}}

\newcommand{\xbf}{\mathbf{x}}

\newcommand{\Xbf}{\mathbf{X}}

\newcommand{\Lcal}{\mathcal{L}}

\newcommand{\Rcal}{\mathcal{R}}

\newcommand{\Vcal}{\mathcal{V}}

\DeclareMathAlphabet{\mathsfit}{\encodingdefault}{\sfdefault}{m}{sl}
\SetMathAlphabet{\mathsfit}{bold}{\encodingdefault}{\sfdefault}{bx}{n}

\newif\ifsubmission
\submissionfalse

\title{Extracting Training Data from Diffusion Language Models via Infilling}

\author{%
  Yihan Wang \thanks{Corresponding author.} \\
  University of Waterloo\\
  \texttt{yihan.wang@uwaterloo.ca} \\
  \And
  N. Asokan \\
  University of Waterloo \\
  KTH Royal Institute of Technology \\
  \texttt{asokan@acm.org} \\
}

\begin{document}

\maketitle

\begin{abstract}
  Memorization in large language models has been studied almost exclusively through prefix-conditioned extraction, a natural choice for autoregressive models. However, diffusion language models (DLMs) can denoise masked tokens at arbitrary positions. Thus, prefix-only probing reveals only one facet of memorization in DLMs and significantly underestimates the risk of training-data extraction. 
  In order to realistically model extractability of training data in DLMs, we introduce \emph{infilling extraction}, a data-extraction protocol parameterized by an arbitrary binary mask that subsumes prefix-only probing and accounts for the bidirectional inductive bias of DLMs. Instantiating it on LLaDA-8B and Dream-7B across five extraction modes, three training pipelines, and three corpora covering verbatim and partial leakage, we find that mask geometry governs extractability: edge-conditioned masks \emph{extract up to three times more} verbatim sequences than prefix-conditioned ones, and bidirectional access opens channels inaccessible in autoregressive models.
  In particular, an adversary who obtains a PII-redacted copy of the training data can use targeted infilling to raise PII recall on fine-tuned DLMs by 2.5–4.5× over prefix-conditioned extraction, substantially narrowing the gap to a scale-matched autoregressive model and, for LLaDA on email addresses, exceeding it under greedy prefix-conditioned extraction.
  Tunable parameters for decoding measurably affect extraction performance, while a follow-up supervised fine-tuning stage does not eliminate the prior memorization.
  Code is available at \href{https://github.com/ssg-research/DLM-Infilling-Extraction}{https://github.com/ssg-research/DLM-Infilling-Extraction}.
\end{abstract}

\section{Introduction}\label{sec:introduction}

By learning statistical patterns over vast and diverse text, large language models (LLMs) exhibit surprising generalization ability, successfully adapting to a wide variety of downstream tasks.
However, this reliance on training data comes at a cost: models do not merely generalize, but also memorize portions of their training data, raising serious concerns about the leakage of personally identifiable information (PII), the reproduction of copyrighted material, and the disclosure of confidential content embedded in training corpora~\cite{carlini2023quantifying,LeeINZECC22,KaramLZS23,LukasSSTWZ23b}. A prominent line of work demonstrating these risks is \textit{data extraction attacks}~\cite{CarliTWJHLRBSEOR21,NasrCHJCICWTL23,YuPLDKHLY23,HayesSCYSNCLC25}, in which an adversary crafts prompts that induce the model to regurgitate sequences from its training set.

While memorization and data extraction attacks on autoregressive language models (ARMs) have been studied extensively~\cite{XiongZPS25}, a fundamentally different paradigm of text generation has recently emerged: diffusion language models (DLMs)~\cite{LiCGS25,GongAZYZLAZBHPK25,NieZYZOHZLWL25,YeXZGWJLK25,XuGKNXLEV24}. Rather than predicting the next token strictly left-to-right, DLMs learn to iteratively denoise corrupted text and can generate multiple tokens at arbitrary positions per step~\cite{AustiJHTv21,LiTGLH22,StrudTADGMGSDSL22,HeSTWHQ23}, offering a promising path toward faster and more flexible language models~\cite{SongZLGXLLYYQFSZHWYJLMZWZ25,GeminiDiffusion2026,LabsKKLVWBLMPEGK25,BieCCDGGGHHLLLLLLLLLMTWWXZZZZZZZZ25,BieCCCCCDFFGGGGGHHJJLLLLLLLLMMPQRTTWWWXYZZZZZZZZZZZZ26}. Despite their growing scale and deployment, memorization characteristics of DLMs and their vulnerability to data extraction remain largely unexplored.

In this work, we close this gap by developing a comprehensive framework for evaluating data extractability in large DLMs. Our central observation is that estimating the risk of training data extraction in DLMs using only prefixes from training data records as prompts~\cite{LuoYLB26} (also known as the \emph{prefix-conditioned protocol}) can significantly underestimate the risk because this protocol, inherited from exploring data extraction in ARMs, captures only one facet of memorization: it asks what the model produces \emph{next}, but not what it would fill in given \emph{arbitrary surrounding context}. We therefore introduce \emph{infilling extraction}, a strictly more general protocol parameterized by a binary mask $\mbf$. The unmasked positions of a training sequence constitute the context revealed to a DLM, which is asked to reconstruct the text corresponding to the masked positions. The training sequence is considered extracted if the reconstruction matches the masked text.

Within this protocol, we instantiate five canonical access modes which we term \textit{prefix}/\textit{suffix}-conditioned, \textit{middle} part-conditioned, \textit{edge}-conditioned, and uniformly \textit{random} geometries, plus a \emph{targeted} variant in which only sensitive entities are masked. We investigate three training pipelines that cover how DLMs are realistically exposed to data, including full fine-tuning (FT), low-rank adaptation (LoRA), and supervised fine-tuning (SFT) under a chat template, and evaluate on three datasets that surface complementary risks: MIMIR Pile for verbatim extraction, the Enron email for PII extraction, and Tulu3 for prompt/response leakage in instruction-tuned deployments. We additionally trace how the practitioner-tunable parameters of decoding (\ie, number of denoising steps, semi-autoregressive block size, sampling temperature, and remasking strategy) affect extractability.

Our experiments on LLaDA-8B~\cite{NieZYZOHZLWL25} and Dream-7B~\cite{YeXZGWJLK25} uncover several patterns that \textit{prefix}-conditioned evaluation alone would have missed. 
(i) Mask geometry governs extractability.  At matched budget, \textit{edge}-conditioned masks \emph{extract up to three times more verbatim sequences} than \textit{prefix}-conditioning, evidencing the bidirectional inductive bias of DLMs.
(ii) Bidirectional access opens new leakage channels. Adversaries can recover user prompts from responses and perform \textit{targeted} PII reconstruction, both invisible under \textit{prefix}-conditioned probing.
(iii) Prefix-only evaluation \cite{LuoYLB26} underestimates the risk of training-data extraction in DLMs. Under prefix-conditioned extraction, DLMs indeed leak less PII than a scale-matched ARM. However, if the adversary obtains a redacted version of the training data, through mandated disclosure (e.g., regulatory or court-ordered release with PII redaction) or via a partial data leak—both of which are realistic scenarios \cite{hhs_deidentification_2012,gdpr,treasury_ai_2026,shaikh2025balancing}---\textit{targeted} raises the PII recall of DLMs by 2.5–4.5× and substantially narrows this gap: on LLaDA-8B, email recall (40.5\%) exceeds that of the ARM under greedy prefix-conditioned extraction (35.9\%), although the ARM remains stronger when it additionally exploits suffix information via sample-then-rank~\cite{LukasSSTWZ23b}.
(iv) Decoding hyperparameters serve as practical amplification or mitigation levers, and a follow-up SFT stage redistributes but does not close the channel that infilling exposes.


We summarize our contributions as follows:
\begin{itemize}
    \item We formalize \emph{infilling extraction} (\Cref{sec:method}), a mask-parameterized data-extraction protocol, generalizing \textit{prefix}-conditioned extraction and accounting for bidirectional bias of DLMs.
    \item We instantiate the protocol with five access modes plus a targeted-PII variant, three training pipelines, and four decoding parameters, yielding the \textit{first systematic study of memorization and training-data extraction in large DLMs} across access patterns, exposure regimes, and inference-time configurations, showing that \emph{DLMs pose a greater risk of training-data leakage than previously thought} (\Cref{subsec:exp:setting}--\ref{subsec:comparison-arm-dlm}).
    \item We empirically characterize how mask geometry, decoding parameters, and post-training stages shape extractability on LLaDA-8B and Dream-7B, \emph{identifying new leakage channels} structurally inaccessible in ARMs and discuss inference-time mitigations (\Cref{subsec:influence-decoding-factors}--\ref{subsec:sft}).
\end{itemize}

\section{Preliminaries}\label{sec:background}
\subsection{Diffusion Language Models}
\label{subsec:dlms}
Whereas ARMs predict the next token strictly from left to right, DLMs cast generation as iterative denoising: a forward process corrupts a sequence by replacing tokens with a special \texttt{[MASK]} symbol, and a learned reverse process recovers them from the partially masked context. Representative masked DLMs such as LLaDA~\cite{NieZYZOHZLWL25} and Dream~\cite{YeXZGWJLK25} are trained with a cross-entropy loss over the masked positions,
\begin{equation*}
    \Lcal(\theta) = -\Ebb_{t,\xbf_0,\mbf_t}\!\left[
    \frac{1}{t}\sum_{i=1}^L \mbf_t^i\cdot \log p_\theta(\xbf_0^i\mid \xbf_0\otimes\mbf_t)
    \right],
\end{equation*}
where $\xbf_0$ is a length-$L$ sequence, $t\!\in\!(0,1]$ is a uniformly sampled corruption level, $\mbf_t$ is a random binary mask with $\Vert\mbf_t\Vert_0=\lfloor t L \rfloor$, $\otimes$ is mask operation and $\xbf_0\otimes\mbf_t$ masks a fraction $t$ of $\xbf_0$. 

At inference, a DLM unmasks multiple positions per step over $T$ denoising steps~\cite{WuZXLDZLHX25,IsraeBG25}; which positions to commit is governed by a \emph{remasking} strategy (random, confidence-based, or uncertainty-based)~\cite{NieZYZOHZLWL25,YeXZGWJLK25}. Semi-autoregressive decoding further partitions the sequence into blocks of size $b$, decoding within each block in parallel while preserving order across blocks~\cite{WuZXLDZLHX25,ArrioGCYQHSK25}, and sampling of temperature $\tau$ can be applied to the per-position distributions. 
Beyond left-to-right prompted generation, the bidirectional formulation enables \emph{infilling}~\cite{ZhangXZLCWG25,FujinS26,WenQLLWYJXLLLSHZ25}: given a sequence with arbitrary masked spans, the model predicts the missing tokens from both preceding and following context (e.g., filling ``Joe Biden'' into ``\texttt{[MASK] [MASK] is the 46th president of the United States}'').

\subsection{Data Extraction}\label{subsec:data-extraction}
In data extraction attacks, an adversary queries a model to elicit sequences appearing in the training set, potentially recovering PII, copyrighted text, or other sensitive content~\cite{ZengLRLXHXWTY24,udsa2025exploring}. The typical data extraction protocol, \textit{prefix}-conditioning, is a natural choice against ARMs: a training sequence $\xbf\in\Vcal^n$ is split into a prefix $\xbf_{\le k}$ and suffix $\xbf_{>k}$, and $\xbf$ is considered extracted if the text predicted by the model, when prompted with the prefix, matches the suffix, $g_\theta(\xbf_{\le k}) = (\xbf_{\le k}, \xbf_{>k})=\xbf$, in the case of a verbatim match, where $g_\theta$ is the decoding function. Memorization is summarized by the extraction rate over training samples, often paired with a verbatim-match criterion such as a contiguous $50$-token span~\cite{CarliTWJHLRBSEOR21,carlini2023quantifying}; probabilistic variants replace the deterministic match with the probability of producing the suffix under stochastic sampling~\cite{HayesSCYSNCLC25}. 


\section{Related Work}\label{sec:related_work}
\textbf{DLM extraction.} Two recent works study data extraction in DLMs along distinct axes. \citet{LuoYLB26} generalize prefix-based $(n,p)$-discoverable extraction~\cite{HayesSCYSNCLC25} to arbitrary masks. Under \emph{prefix}-conditioned PII extraction, they report that DLMs leak less than scale-matched ARMs. \citet{ChenZDSFKRL26} instead target membership inference attacks (MIAs), exploiting multiple maskable configurations of DLMs via progressively densified masks and sign-based, inverse-step-weighted aggregation to substantially improve the effectiveness of MIAs on fine-tuned LLaDA-8B and Dream-7B.
Our work differs in both the problem addressed and the extraction protocol used. Our threat model is the extraction of verbatim training data or PIIs present in training data, a strictly stronger criterion than the binary membership decision of \citet{ChenZDSFKRL26}. 
Although \citet{LuoYLB26} generalizes prefix-based extraction to arbitrary masks, their PII extraction experiments revert to the prefix-only setting when comparing DLMs and ARMs.
We argue that it artificially restricts an adversary's capabilities in the case of DLMs and therefore underestimates the risk of data leakage. We parameterize extraction by an arbitrary mask $\mbf$ and instantiate it across five access modes, three training pipelines, and four decoding parameters. Our results illuminate the findings of \citet{LuoYLB26}: although we confirm their observation that prefix-conditioned leakage appears modest, we show that alternative mask patterns result in substantially more leakage.

\textbf{ARM extraction with suffix information.} \citet{LukasSSTWZ23b} proposed a "sample-then-rank" attack against ARMs: PII candidates are sampled conditioned on the prefix, the suffix is then appended, and the final candidate is selected by perplexity ranking. Though ARMs can't use suffix natively, this attack exploits suffix information to mount a stronger PII extraction attack against ARMs. 
Our work, by contrast, establishes a unified framework for both verbatim and PII extraction against DLMs, whose bidirectional nature naturally supports a much wider range of infilling capabilities.

\textbf{BERT Extraction.} Prior work \cite{lehman-etal-2021-bert,vakili2021clinical,diera2023memorization} examined extraction attacks on BERT~\cite{devlin2019bert} over clinical and general-domain data, and all report negative results: infilling could not reliably extract PII entities from finetuned BERT models. \citet{xia2025exploring} studied probe sensitivity for BERT-based named entity recognition models, which does not constitute a data extraction threat model. \citet{carlini2023quantifying,tirumala2022memorization} examined fill-in-the-blank extraction on BERT and T5~\cite{raffel2020exploring}, but are largely restricted to random, single-token completion and do not address PII extraction.
Our work differs from prior BERT extraction efforts in object of study, framework, task scope, and even qualitative conclusions. 
We show that infilling extraction attacks are effective against DLMs — the emerging next generation of masked language models — and we further study how memorization manifests under different infilling protocols and how decoding parameters affect extraction.

\textbf{Image Diffusion Extraction.} Carlini et al.~\cite{carlini2023extracting} show that image diffusion models such as Stable Diffusion memorize and regenerate individual training images. They also propose an inpainting attack, in which masked regions of training images are reconstructed more faithfully than those of non-members. In this work, we compare multiple mask geometries on discrete text, extract PII, and analyze decoding parameters specific to DLMs.
\section{Infilling Extraction}\label{sec:method}
Memorization in DLMs can be exploited by providing partial context anywhere in a sequence; prefix-only conditioning, therefore, is just one possible approach. DLMs natively support \emph{infilling}, in which any subset of tokens may be masked and reconstructed. This enables a strictly more general extraction protocol than prefix-only conditioning.

\begin{definition}[Infilling Extraction]
Let $\theta$ denote a DLM, $\xbf \in \Vcal^n$ a training sequence, $\mbf \in \{0,1\}^n$ a binary mask, and $g_\theta$ the decoding function. We say $\xbf$ is \emph{verbatim extracted} under infilling if
\begin{align} \label{eq:infilling}
    g_\theta(\xbf \otimes \mbf) = \xbf.
\end{align}
\end{definition}
The mask $\mbf$ specifies the access pattern (which tokens are revealed), while $g_\theta$ instantiates the inference procedure characterized in \Cref{subsec:dlms} along four parameters $\{T, b, \tau, \text{remask strategy}\}$. We discuss the effect of possible access patterns (\Cref{subsec:extraction-modes}) and decoding parameters of $g_\theta$ (\Cref{subsec:influence-decoding-factors}).

\subsection{Extraction Modes}\label{subsec:extraction-modes}
Infilling extraction is parameterized by the binary mask $\mbf\in\{0,1\}^n$: positions with $\mbf_i=1$ are masked and to be reconstructed, while the unmasked positions constitute the context revealed to the model. Different patterns of $\mbf$ probe different forms of extractability; for instance, under a fixed context budget, does the same training sequence remain extractable when the revealed context is moved from the beginning to the end of the sequence? To make this comparison concrete and to span the access patterns enabled by bidirectional decoding, we fix a budget of $k$ unmasked tokens (out of $n$) and define five canonical modes:

\begin{itemize}[leftmargin=20pt] 
    \item \textit{Prefix-conditioned}: the first $k$ tokens are revealed and the remaining $n-k$ are masked. This recovers the de facto setting of prior ARM extraction work and serves as our reference.

    \item \textit{Suffix-conditioned}: only the last $k$ tokens are revealed, with the preceding $n-k$ masked. This probes reconstruction in reverse contextual order; under the chat template, it corresponds to recovering a prompt (conversation history) from a known response.

    \item \textit{Edge-conditioned}: the first $k/2$ and last $k/2$ tokens are revealed, and the middle is masked, testing whether bracketing context on both sides amplifies recall relative to one-sided context of the same total budget.

    \item \textit{Middle-conditioned}: a contiguous span of $k$ tokens in the middle is revealed, and both ends are masked, which is complementary to \textit{edge}. This mode isolates whether internal context alone can elicit memorized continuations.

    \item \textit{Random}: a uniformly random subset of $k$ positions is revealed, and the remaining $n-k$ are masked, probing recovery from arbitrarily distributed context and reflecting diffuse, non-localized memorization.
\end{itemize}

Beyond these canonical budget-constrained modes, the PII experiments in \Cref{sec:exp} additionally consider a \emph{targeted} infilling setting, in which only the tokens of a sensitive entity are masked, and the entire surrounding sequence is revealed as context. 
Targeted infilling probes entity-level rather than sequence-level memorization and characterizes an upper bound on PII extractability under a maximally informed adversary.

\subsection{Extraction Task and Metrics}\label{subsec:extraction-task-metrics}

We evaluate infilling extraction attacks on two tasks: \textit{verbatim extraction} and \textit{PII extraction}. The former measures sequence-level memorization, i.e., the extent to which the model reproduces specific training samples; the latter measures entity-level memorization of PII contained in the training data.

For verbatim extraction, we sample a set $\Xbf$ of snippets from the training dataset and mask them according to a predefined context access and budget. We then perform conditional or \textit{random} infilling based on the mask pattern. We compute the attack success rate (ASR) as the proportion of samples extracted verbatim:
\begin{align*}
    \textbf{ASR}(\theta,\Xbf) \coloneqq \frac{1}{|\Xbf|}\sum_{\xbf \in \Xbf} \one[g_\theta (\xbf \otimes \mbf) = \xbf ].
\end{align*}
Since ASR is a strict metric that captures only exact extraction, we further relax the criterion by introducing the token-wise extraction rate (TER), which measures the fraction of correctly recovered tokens at masked positions:
\begin{align*}
    \textbf{TER}(\theta,\Xbf) \coloneqq \frac{1}{|\Xbf|}\sum_{\xbf \in \Xbf}  \frac{1}{\Vert \mbf\Vert_0}  \sum_{i=1}^n \mbf_i\cdot \one[g_\theta (\xbf \otimes \mbf)_i = \xbf_i].
\end{align*}

For PII extraction, we prompt the model with an unmasked \textit{prefix}, \textit{suffix}, \textit{edge}, \textit{middle} context, as well as conduct \textit{targeted} infilling.
For conditional extraction, we apply a function $\Rcal$ (\eg, regex rule) to extract PII entities from both the generated text and the ground truth, and compute recall and precision as follows:
\begin{align*}
    \textbf{Recall}(\theta,\Xbf) &\coloneqq 
    \frac{\left|\Rcal\!\left(\{g_\theta(\xbf\otimes\mbf)\mid\xbf\in\Xbf\}\right)
    \cap
    \Rcal(\Xbf)\right|}{|\Rcal(\Xbf)|}, \\
    \textbf{Precision}(\theta,\Xbf) &\coloneqq 
    \frac{\left|\Rcal\!\left(\{g_\theta(\xbf\otimes\mbf)\mid\xbf\in\Xbf\}\right)
    \cap
    \Rcal(\Xbf)\right|}{\left|\Rcal\!\left(\{g_\theta(\xbf\otimes\mbf)\mid\xbf\in\Xbf\}\right)\right|}.
\end{align*}
For \textit{targeted} infilling, we report the TER, recall of exactly matched PII spans, and the error-tolerant variants of recall.

\subsection{Training-Induced Memorization}\label{subsec:training-pipelines}
As it is infeasible to access the accurate pre-training data of existing large DLMs like LLaDA and Dream, we instead induce memorization via fine-tuning and extract the data used in fine-tuning. To cover the regimes under which DLMs are realistically exposed to training data, we instantiate the infilling-extraction protocol on three pipelines~\cite{ZhouCTS26}.

First, FT updates all model parameters on the target corpus under the masked-denoising objective of the underlying DLM. It simulates the continual pretraining process.
Second, LoRA~\cite{HuSWALWWC21} freezes the base model and trains a small set of low-rank update matrices, mirroring how DLMs are most likely to be specialized in production. The denoising objective is unchanged. 
Last, SFT trains the model on (prompt, response) pairs wrapped in the chat template, with the loss applied over the full templated sequence so that both prompt and response tokens participate in the denoising objective. This pipeline targets aligned, deployment-facing DLMs and enables the prompt-/response-conditioned extraction modes used to probe whether private user prompts can be recovered from known model responses.

\subsection{Research Questions}\label{subsec:evaluation-goal}
The infilling extraction framework enables a systematic investigation of memorization in DLMs. We pose the following research questions and structure our experiments in \Cref{sec:exp} around them:

\textit{RQ1: How does memorization manifest in DLMs under the infilling-extraction protocol?} 
By applying the infilling extraction protocol across different mask patterns, we can characterize how the placement of context affects the model's ability to recall training data (\Cref{subsec:verbatim-extraction,subsec:pii-extraction}). 
We also analyze the influence of post-training adaptation on the prior memorization, testing whether downstream alignment attenuates pre-existing leakage (\Cref{subsec:sft}).

\textit{RQ2: What risks does extractable memorization in DLMs entail?} Beyond quantifying the extent of memorization, we evaluate the nature of the extracted content to assess the practical risks. We analyze whether the extracted sequences contain PII (\Cref{subsec:pii-extraction}). 
Moreover, in the context of SFT, we test whether user prompts can be extracted from known model responses, which would indicate a risk of prompt leakage in deployed systems (\Cref{subsec:verbatim-extraction}).

\textit{RQ3: Do DLMs pose lower data-leakage risk than ARMs in realistic settings?}  \cite{LuoYLB26} reported that DLMs have a lower risk of training data leakage than scale-matched ARMs under prefix-conditioned extraction. 
We provide a more nuanced understanding of their findings: under a realistic scenario where the adversary has redacted training samples, DLMs can use \textit{targeted} infilling to recover PII (\Cref{subsec:comparison-arm-dlm}). 

\textit{RQ4: How do decoding-time parameters modulate extraction?} The decoding parameters of $g_\theta$ (see \Cref{subsec:dlms}) are routinely tuned by practitioners for speed and quality. We test which of them materially modulate extraction so that the same factor can act as inference-time amplifications or mitigations (\Cref{subsec:influence-decoding-factors}).
Additionally, we trace the per-step trajectory of recovered content to test whether memorized tokens are committed early (the model effectively already ``knows'' the answer) or surface only near the end of denoising (\Cref{subsec:extraction-trajectory}).

\section{Experiments}\label{sec:exp}

\subsection{Setup}\label{subsec:exp:setting}

\textbf{Datasets.}
We instantiate the infilling-extraction protocol on three datasets, covering verbatim extraction, PII extraction, and prompt/response extraction. Following \citet{ChenZDSFKRL26}, we fine-tune on the \textit{full Pile} subset of MIMIR~\citep{DuanSMMSZTCEH24a} ($10{,}000$ sequences) and evaluate verbatim extraction on uniformly sampled $100$-token snippets from the training set. 
We set the context budget $k=50$ tokens.
For PII extraction, we fine-tune on $10{,}000$ messages from the Enron Email corpus~\cite{klimt2004enron}. For each extraction, we sample $100$-token snippets whose masked region contains target entities; we report results on two PII types, \textit{email address} and \textit{phone number}.
For SFT, we fine-tune on $10{,}000$ (prompt, response) pairs from Tulu3~\cite{LambeMPHIBMLDLGMGHYBTWSSWDH25} and restrict evaluation to records whose prompt and response are each at most $100$ tokens.

\textbf{Models.}
We build on the open-source diffusion-LM framework of \citet{ZhouCTS26}, which unifies training, inference, and evaluation across recent DLMs. We use the pre-trained LLaDA-8B-Base~\cite{NieZYZOHZLWL25} and Dream-7B-Base~\cite{YeXZGWJLK25} as DLM backbones, and LLaMA-2-7B~\cite{touvron2023llama} as ARM baseline at matched scale. To ensure measurable memorization, all models are fine-tuned for $5$ epochs on every task; full training details are deferred to \Cref{app-subsec:models}. By default, we use greedy sampling and decode one token per step; LLaDA uses low-confidence remasking with no semi-autoregressive blocking, and Dream uses entropy-based remasking. We vary each of these parameters in \Cref{subsec:influence-decoding-factors}.

\textbf{Infilling extraction.}
On MIMIR Pile we evaluate all five extraction modes (\textit{prefix}, \textit{suffix}, \textit{edge}, \textit{middle}, \textit{random}). On Enron, we evaluate \textit{prefix}, \textit{suffix}, \textit{edge}, \textit{middle}, and introduce \textit{targeted} infilling. We use the Presidio analyzer~\cite{MsPresidio} for email address recognition and a regex rule for phone numbers as the extraction function $\Rcal$ in \Cref{subsec:extraction-task-metrics}. On Tulu3, we evaluate two SFT-specific variants under the chat template, \textit{prompt-conditioned} and \textit{response-conditioned} extraction, in which the response (resp.\ prompt) is masked given the other side as context.

\begin{table}[htbp!]
\centering
\caption{
Verbatim extraction (\%) under different infilling extraction modes on MIMIR Pile. We report ASR and TER averaged over five models trained with different random seeds, along with the standard deviation. \textbf{Bold} denotes the best score per metric, and we \colorbox{yellow!20}{highlight} cases where the best result exceeds \textit{prefix}'s ASR by at least $2\times$.
}
\label{tab:mimir}
\resizebox{\textwidth}{!}{%
\begin{tabular}{l|ccccc|ccccc}
\toprule\midrule
\multirow{2}{*}{Model} & \multicolumn{5}{c|}{ASR $\uparrow$} & \multicolumn{5}{c}{TER $\uparrow$} \\ \cmidrule(lr){2-6} \cmidrule(lr){7-11}
 & \multicolumn{1}{c}{Prefix} & \multicolumn{1}{c}{Suffix} & \multicolumn{1}{c}{Edge} & \multicolumn{1}{c}{Middle} & \multicolumn{1}{c|}{Random} & \multicolumn{1}{c}{Prefix} & \multicolumn{1}{c}{Suffix} & \multicolumn{1}{c}{Edge} & \multicolumn{1}{c}{Middle} & \multicolumn{1}{c}{Random} \\ \midrule
LLaDA-8B-FT & \colorbox{yellow!20}{0.18}{\tiny $\pm$0.02} & 0.17{\tiny $\pm$0.05} & \colorbox{yellow!20}{\textbf{0.41}}{\tiny $\pm$0.10} & 0.12{\tiny $\pm$0.03} & {0.37}{\tiny $\pm$0.09} & 5.14{\tiny $\pm$0.23} & 5.28{\tiny $\pm$0.18} & {8.28}{\tiny $\pm$0.29} & 8.08{\tiny $\pm$0.22} & \textbf{56.22}{\tiny $\pm$1.11} \\
Dream-7B-FT & \colorbox{yellow!20}{0.04}{\tiny $\pm$0.04} & 0.00{\tiny $\pm$0.00} & \colorbox{yellow!20}{\textbf{0.14}}{\tiny $\pm$0.07} & 0.02{\tiny $\pm$0.01} & {0.11}{\tiny $\pm$0.07} & 4.37{\tiny $\pm$0.57} & 4.18{\tiny $\pm$0.42} & 6.93{\tiny $\pm$0.96} & {6.99}{\tiny $\pm$0.92} & \textbf{55.85}{\tiny $\pm$7.34} \\
LLaDA-8B-LoRA & \colorbox{yellow!20}{0.29}{\tiny $\pm$0.01} & 0.42{\tiny $\pm$0.02} & \colorbox{yellow!20}{\textbf{0.72}}{\tiny $\pm$0.04} & 0.25{\tiny $\pm$0.01} & {0.69}{\tiny $\pm$0.04} & 5.97{\tiny $\pm$0.03} & 6.18{\tiny $\pm$0.06} & {9.46}{\tiny $\pm$0.05} & 9.08{\tiny $\pm$0.05} & \textbf{58.52}{\tiny $\pm$0.03} \\
Dream-7B-LoRA & \colorbox{yellow!20}{0.17}{\tiny $\pm$0.02} & 0.00{\tiny $\pm$0.00} & \colorbox{yellow!20}{\textbf{0.34}}{\tiny $\pm$0.03} & 0.02{\tiny $\pm$0.00} & {0.20}{\tiny $\pm$0.03} & 5.27{\tiny $\pm$0.06} & 5.14{\tiny $\pm$0.04} & {8.43}{\tiny $\pm$0.03} & 8.29{\tiny $\pm$0.06} & \textbf{58.61}{\tiny $\pm$0.06} \\
\midrule\bottomrule
\end{tabular}%
}
\end{table}

\subsection{Verbatim Extraction}
\label{subsec:verbatim-extraction}
We evaluate the five infilling modes on MIMIR Pile and the two SFT-specific modes on Tulu3.

\textbf{Mask geometry shapes verbatim extraction.}
Under a fixed budget of $k$ revealed tokens, ASR varies sharply with where those tokens sit. On MIMIR Pile (\Cref{tab:mimir}), LLaDA-8B-FT peaks at \textit{edge} ($0.41\%$) mode, that is, over $2\times$ the one-sided \textit{prefix} ($0.18\%$); the \textit{edge}/\textit{random} $>$ \textit{prefix}/\textit{suffix}/\textit{middle} ordering in ASR reproduces on Dream-7B and under LoRA across both backbones. 
Verbatim extraction is thus governed largely by whether that context spans the structural endpoints of the training sequence.
It confirms the bidirectional inductive bias of DLMs.

\begin{wraptable}[6]{R}{0.5\textwidth}
\centering
\vspace{-12pt}
\caption{Verbatim extraction (\%) on Tulu3.}
\vspace{-5pt}
\label{tab:tulu-sft-results}
\resizebox{0.5\textwidth}{!}{%
\begin{tabular}{l|cc|cc}
\toprule \midrule
\multirow{2}{*}{Model} & \multicolumn{2}{c|}{ASR $\uparrow$} & \multicolumn{2}{c}{TER $\uparrow$} \\
\cmidrule(lr){2-3} \cmidrule(lr){4-5}
 & Prompt & Response & Prompt & Response \\
\midrule
LLaDA-8B-SFT & \textbf{10.32} {\scriptsize $\pm$0.57} & \phantom{0}3.36 {\scriptsize $\pm$1.98} & \textbf{20.69} {\scriptsize $\pm$0.58} & 11.01 {\scriptsize $\pm$1.25} \\
Dream-7B-SFT & \textbf{10.62} {\scriptsize $\pm$0.86} & \phantom{0}1.78 {\scriptsize $\pm$0.24} & \textbf{20.16} {\scriptsize $\pm$0.29} & \phantom{0}9.56 {\scriptsize $\pm$0.43} \\
\midrule \bottomrule
\end{tabular}%
}
\end{wraptable}
\textbf{Token-level recovery does not imply verbatim extraction.}
\Cref{tab:mimir} reveals a striking disparity between TER and ASR: under \textit{random} masking, LLaDA-8B-FT recovers $56.22\%$ of masked tokens yet reproduces only $0.37\%$ of snippets verbatim, with Dream-7B-FT exhibiting an even sharper gap ($55.85\%$ vs.\ $0.11\%$). This high token accuracy is largely an artifact of the scaffolding provided by the revealed tokens, and does not translate into successful verbatim extraction. We therefore interpret TER as a loose indicator of memorization risk and adopt ASR as the operational threat metric. 

\textbf{Prompts are recoverable from responses.}
On Tulu3, the standard \textit{prompt-conditioned} direction reaches ASR $10.32\%$/$10.62\%$ for LLaDA-/Dream-SFT, but the reverse \textit{response-conditioned} direction, where the response is revealed and prompt is to be inferred, also remains non-trivial (ASR $3.36\%$/$1.78\%$). This leakage channel is structurally inaccessible to ARMs: bidirectional denoising allows an adversary holding a deployed model's output to recover the originating user query, with direct implications for instruction-tuned DLMs queried with sensitive prompts.

\begin{table}[htbp!]
\centering
\caption{PII extraction (\%) under different infilling extraction modes on Enron Email dataset. We \colorbox{yellow!20}{highlight} cases where the best result exceeds \textit{prefix}'s recall by at least $2\times$.
}
\label{tab:pii_results}
\resizebox{\textwidth}{!}{
\begin{tabular}{l|l|cccc|cccc}
\toprule\midrule
\multirow{2}{*}{PII Type} & \multicolumn{1}{c|}{\multirow{2}{*}{Model}} & \multicolumn{4}{c|}{Recall $\uparrow$} & \multicolumn{4}{c}{Precision $\uparrow$} \\
\cmidrule(lr){3-6} \cmidrule(lr){7-10}
 & & Prefix & Suffix & Edge & Middle & Prefix & Suffix & Edge & Middle \\
\midrule
\multirow{4}{*}{\shortstack[l]{Email\\Address}}
 & LLaDA-8B-FT    & \colorbox{yellow!20}{1.95}{\tiny$\pm$0.32} & \phantom{0}2.20{\tiny$\pm$0.41} & \colorbox{yellow!20}{\textbf{6.33}}{\tiny$\pm$1.42} & \phantom{0}{2.71}{\tiny$\pm$0.62} & \textbf{33.71}{\tiny$\pm$11.10} & 21.94{\tiny$\pm$4.49} & {27.92}{\tiny$\pm$1.70} & 19.14{\tiny$\pm$2.92} \\
 & Dream-7B-FT    & \colorbox{yellow!20}{0.26}{\tiny$\pm$0.55} & \phantom{0}0.30{\tiny$\pm$0.64} & \colorbox{yellow!20}{\textbf{0.93}}{\tiny$\pm$1.97} & \phantom{0}{0.46}{\tiny$\pm$1.02} & \phantom{0}{3.24}{\tiny$\pm$6.16} & \phantom{0}3.22{\tiny$\pm$6.86} & \phantom{0}\textbf{4.08}{\tiny$\pm$8.13} & \phantom{0}2.81{\tiny$\pm$6.27} \\
 & LLaDA-8B-LoRA  & \colorbox{yellow!20}{0.51}{\tiny$\pm$0.17} & \phantom{0}1.29{\tiny$\pm$0.55} & \colorbox{yellow!20}{\textbf{4.55}}{\tiny$\pm$0.75} & \phantom{0}{1.44}{\tiny$\pm$0.37} & {16.62}{\tiny$\pm$4.75} & {16.62}{\tiny$\pm$5.80} & \textbf{26.72}{\tiny$\pm$2.60} & 15.75{\tiny$\pm$2.24} \\
 & Dream-7B-LoRA  & \colorbox{yellow!20}{0.41}{\tiny$\pm$0.13} & \phantom{0}0.80{\tiny$\pm$0.34} & \colorbox{yellow!20}{\textbf{2.70}}{\tiny$\pm$0.29} & \phantom{0}{1.40}{\tiny$\pm$0.36} & \phantom{0}4.43{\tiny$\pm$0.73} & 11.25{\tiny$\pm$5.03} & \textbf{14.26}{\tiny$\pm$1.40} & {12.43}{\tiny$\pm$2.31} \\
\midrule
\multirow{4}{*}{\shortstack[l]{Phone\\Number}}
 & LLaDA-8B-FT    & \colorbox{yellow!20}{1.75}{\tiny$\pm$1.28} & \phantom{0}1.43{\tiny$\pm$0.51} & \phantom{0}{4.56}{\tiny$\pm$2.71} & \colorbox{yellow!20}{\textbf{4.56}}{\tiny$\pm$2.28} & 10.46{\tiny$\pm$6.40} & 16.18{\tiny$\pm$6.58} & {18.26}{\tiny$\pm$8.21} & \textbf{23.38}{\tiny$\pm$10.86} \\[1.8pt]
 & Dream-7B-FT    & \phantom{0}3.31{\tiny$\pm$1.48} & \phantom{0}1.31{\tiny$\pm$0.90} & \phantom{0}{\textbf{3.81}}{\tiny$\pm$2.06} & \phantom{0}{3.53}{\tiny$\pm$1.64} & 14.88{\tiny$\pm$4.45} & 16.20{\tiny$\pm$9.48} & {16.73}{\tiny$\pm$6.49} & \textbf{18.82}{\tiny$\pm$7.86} \\
 & LLaDA-8B-LoRA  & \colorbox{yellow!20}{1.01}{\tiny$\pm$0.48} & \phantom{0}0.72{\tiny$\pm$0.27} & \phantom{0}{2.24}{\tiny$\pm$0.66} & \colorbox{yellow!20}{\textbf{2.54}}{\tiny$\pm$0.74} & \phantom{0}5.93{\tiny$\pm$2.64} & {10.23}{\tiny$\pm$3.98} & \phantom{0}9.36{\tiny$\pm$2.60} & \textbf{15.47}{\tiny$\pm$4.20} \\
 & Dream-7B-LoRA  & \phantom{0}2.72{\tiny$\pm$0.67} & \phantom{0}1.63{\tiny$\pm$0.47} & \phantom{0}{3.50}{\tiny$\pm$1.11} & \phantom{0}{\textbf{3.53}}{\tiny$\pm$0.33} & \phantom{0}8.01{\tiny$\pm$1.84} & {15.87}{\tiny$\pm$6.19} & \phantom{0}9.97{\tiny$\pm$2.44} & \textbf{16.95}{\tiny$\pm$3.60} \\
\midrule\bottomrule
\end{tabular}}
\end{table}
\subsection{PII Extraction}
\label{subsec:pii-extraction}
We next ask whether infilling extraction can recover PII from the Enron Email training data. \Cref{tab:pii_results} reports recall and precision for email addresses and phone numbers under the four conditional modes. \Cref{tab:targeted-infilling} probes the worst case in which only the PII tokens are masked, which serves as an \textit{upper bound} for extraction performance.

\textbf{Bidirectional context amplifies PII leakage.}
As shown in \Cref{tab:pii_results}, the \textit{edge} and \textit{middle} modes, which exploit bidirectional context, consistently extract more PII entities than the one-sided \textit{prefix} and \textit{suffix} modes. On LLaDA-8B-FT, \textit{edge} achieves $6.33\%$ recall on emails, 
that is, around $3\times$ that of \textit{prefix} ($1.95\%$) and \textit{suffix} ($2.20\%$) under a matched budget. In terms of extraction precision, \textit{edge} and \textit{middle} likewise outperform their one-sided counterparts in most cases.

\begin{table}[ht]
    \centering
    \caption{\textit{Targeted} infilling results (\%) on Enron dataset. We report TER and error-tolerant recall metrics. Recall$_k$ denotes the fraction of masked entities for which at most $k$ tokens are incorrectly recovered.}
    \label{tab:targeted-infilling}
    \resizebox{0.8\textwidth}{!}{
    \begin{tabular}{l|cccc|cccc}
        \toprule\midrule
        \multirow{2}{*}{Model} & \multicolumn{4}{c|}{Email Address} & \multicolumn{4}{c}{Phone Number} \\
        \cmidrule(lr){2-5} \cmidrule(lr){6-9}
         & TER & Recall & Recall$_1$ & Recall$_2$ & TER & Recall & Recall$_1$ & Recall$_2$ \\
        \midrule
       LLaDA-8B-Base & 14.2 & \phantom{0}1.4 & \phantom{0}2.1 & \phantom{0}5.2 & 38.5 & \phantom{0}0.3 & \phantom{0}0.3 & \phantom{0}0.7 \\
        Dream-7B-Base & 25.4 & \phantom{0}2.1 & \phantom{0}4.4 & \phantom{0}9.3 & 42.2 & \phantom{0}0.1 & \phantom{0}1.0 & \phantom{0}3.3 \\
        \midrule
        LLaDA-8B-FT   & 52.9 & 40.5 & 43.4 & 47.3 & 63.3 & 28.9 & 35.3 & 41.3 \\
        Dream-7B-FT   & 38.7 & 10.8 & 18.6 & 27.0 & 67.2 & 35.8 & 42.0 & 47.9 \\
        \midrule\bottomrule
        \end{tabular}%
        }
\end{table}

\textbf{Targeted infilling.}
To capture the strongest data leakage risk in realistic deployment scenarios, we consider the following threat model. The model trainer (\eg, companies, government agencies, hospitals) has full access to sensitive data and uses it to train a DLM. The adversary later obtains a redacted version of the training data, either through partial leakage or because the trainer was compelled (e.g., by regulation or court order) to release the training data with PII redacted~\cite{hhs_deidentification_2012,gdpr,treasury_ai_2026,shaikh2025balancing}. The adversary's goal is to recover the redacted content by exploiting the model. Specifically, the adversary aligns masks with the redacted positions and queries the model to fill in the original PII, using the surrounding unredacted text as context.

\Cref{tab:targeted-infilling} reports TER and recall$_k$, the fraction of entities recovered with at most $k$ mistaken tokens. Two findings stand out. (i) Fine-tuning sharply increases extractability: on LLaDA-8B, exact-match recall rises from $1.4\%$ to $40.5\%$ for emails and from $0.3\%$ to $28.9\%$ for phone numbers, with the same trend on Dream-7B. This confirms that the recovered entities are genuine training strings rather than distributional guesses. (ii) Recall$_2$ exceeds exact-match recall by $6.8$ points for emails and $12.4$ points for phone numbers. Even partial recoveries can leak meaningful information, and a small number of mistaken tokens is often enough to seed enumeration attacks. Therefore, \textit{targeted} infilling characterizes the upper-bound risk surface for sensitive entities embedded in DLM training data.

\begin{figure}
    \centering
    \includegraphics[width=0.7\linewidth]{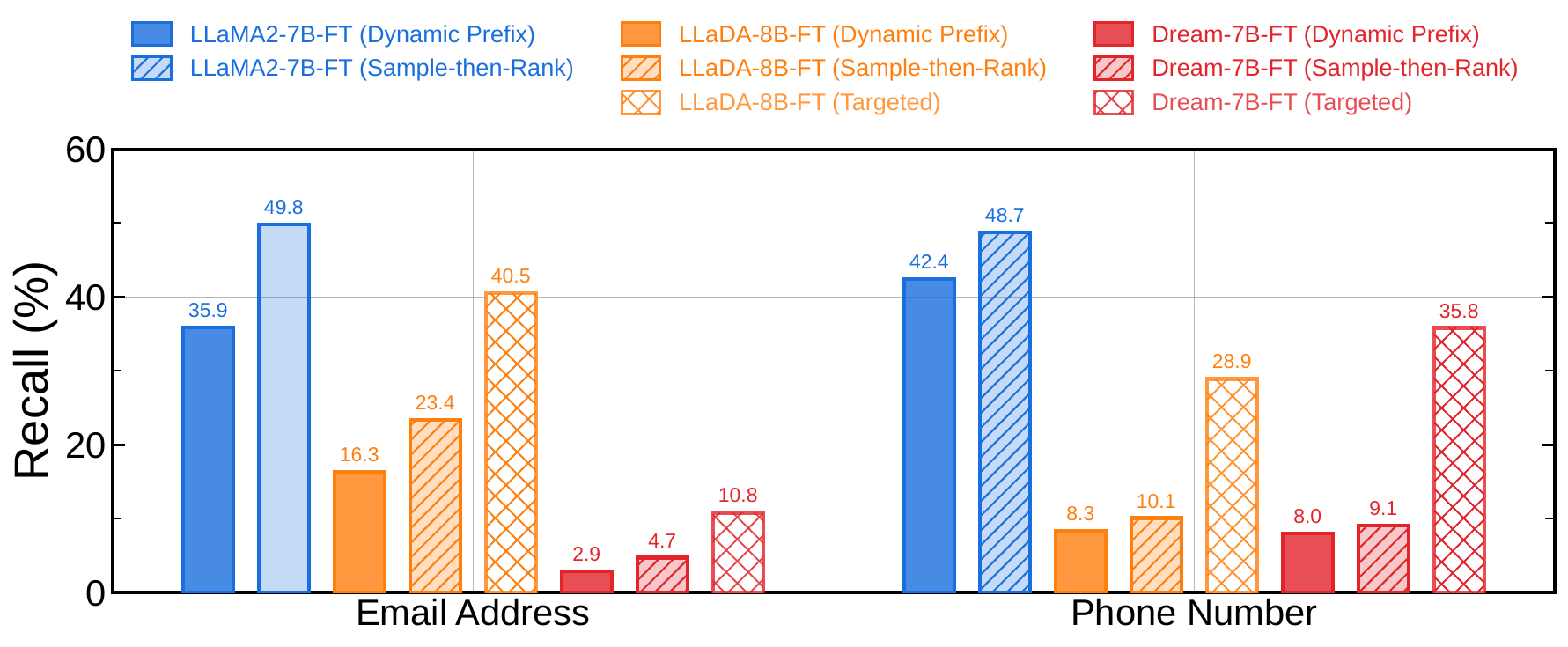}
        \caption{Comparison between ARM and DLMs on Enron.}
        \label{fig:compare-ar-dlm-target}
        \vspace{-20pt}
\end{figure}

\subsection{Comparison between ARMs and DLMs}
\label{subsec:comparison-arm-dlm}

As ARMs and DLMs differ in architecture and training recipe, a strictly fair comparison between them is hard. As an exploratory study, we compare LLaDA-8B and Dream-7B against LLaMA2-7B under matched fine-tuning on Enron, using three settings: \textit{dynamic prefix} and \textit{sample-then-rank}~\cite{LukasSSTWZ23b} for both the ARM and the DLMs, and \textit{targeted} infilling for the DLMs.
In the \textit{dynamic prefix} setting, the revealed context is truncated immediately before the first PII token, which differs from the fixed-budget \textit{prefix} in \Cref{tab:pii_results} and is the strongest \textit{prefix}-conditioned extraction against ARMs under greedy decoding. In the \textit{sample-then-rank} setting, we sample 16 candidates at temperature 1.0 from the dynamic prefix, append the suffix to each, and select the one with the lowest perplexity.

\Cref{fig:compare-ar-dlm-target} shows the comparison.
Under \textit{dynamic prefix}, DLMs leak substantially less than the ARM, consistent with \citet{LuoYLB26}: LLaMA2 attains $35.9\%$ recall on email addresses, roughly $2\times$ that of LLaDA ($16.3\%$) and $12\times$ that of Dream ($2.9\%$).
\textit{Sample-then-rank} improves email recall by $14.0$, $7.1$, and $1.8$ percentage points for LLaMA2, LLaDA, and Dream, respectively.
In contrast, \textit{targeted} infilling raises DLM recall $2.5$--$4.5\times$ above dynamic prefix: LLaDA reaches $40.5\%$ on email addresses, exceeding the ARM under greedy dynamic prefix, and LLaDA/Dream reach $28.9\%$/$35.8\%$ on phone numbers.
Although the ARM under \textit{sample-then-rank} remains stronger, targeted infilling substantially narrows the gap, revealing that \textit{prefix}-conditioned extraction underestimates the data leakage risks of DLMs in realistic scenarios.

\subsection{Influence of Decoding Parameters}\label{subsec:influence-decoding-factors}
\begin{figure}[htbp!]
        \centering
        \begin{subfigure}[t]{0.24\linewidth}
            \includegraphics[width=\linewidth]{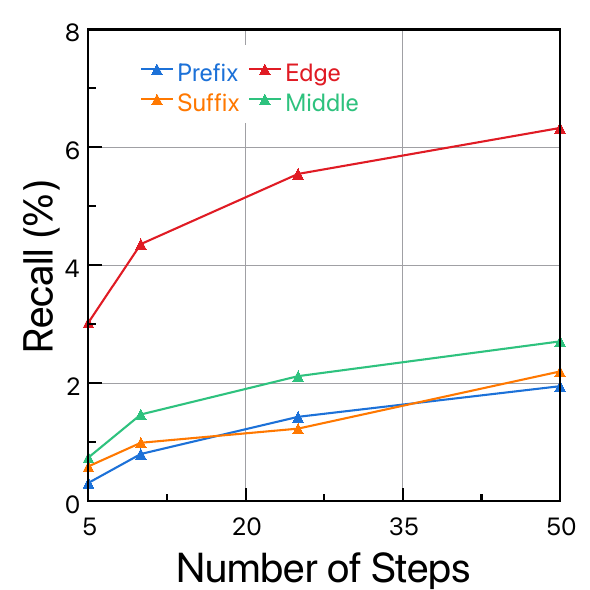}
            \caption{Number of steps $T$}
            \label{fig:enron-step}
        \end{subfigure}
        \hfill
        \begin{subfigure}[t]{0.24\linewidth}
            \includegraphics[width=\linewidth]{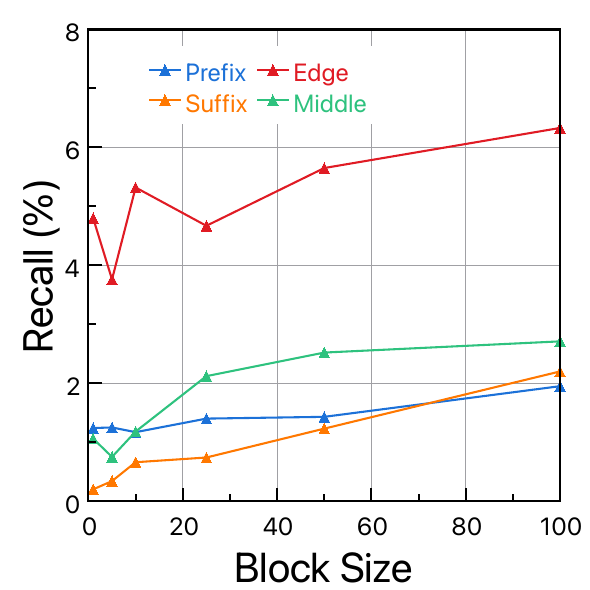}
            \caption{Block size $b$}
        \label{fig:enron-block-size}
        \end{subfigure}
        \hfill
        \begin{subfigure}[t]{0.24\linewidth}
            \includegraphics[width=\linewidth]{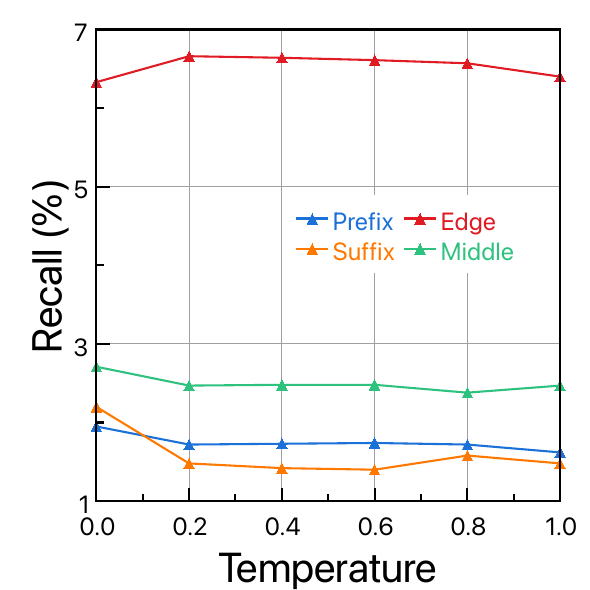}
            \caption{Temperature $\tau$}
        \label{fig:enron-temperature}
        \end{subfigure}
        \hfill
        \begin{subfigure}[t]{0.24\linewidth}
            \includegraphics[width=\linewidth]{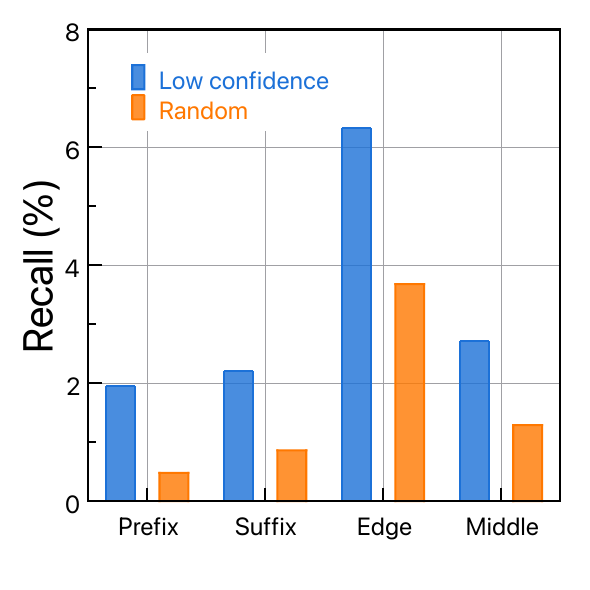}
            \caption{Remasking strategy}
        \label{fig:enron-remasking}
        \end{subfigure}
        \caption{Influence of decoding parameters for LLaDA-8B-FT model on extracting email addresses from Enron dataset.}
        \label{fig:influence-decoding-factors}
\end{figure}

The generation function $g_\theta$ of a DLM exposes several parameters, \ie, denoising steps $T$, semi-autoregressive block size $b$, sampling temperature $\tau$, and remasking strategy, that can be tuned for speed and quality. We separate these decoding parameters and investigate the influence of each component. \Cref{fig:influence-decoding-factors} sweeps each on LLaDA-8B-FT for Enron emails.

\textbf{Decoding efficiency confines extraction efficacy.}
A small number of denoising steps $T$ commits many positions in parallel per step under coarse posteriors, while a large $T$ spreads the trajectory over more refinement rounds. \Cref{fig:enron-step} shows recall rises monotonically with $T$ on Enron emails. Thus, throughput-oriented configurations (small $T$) obscure memorized content under coarse-grained commits, while the accuracy-oriented settings expose the most training data.

\textbf{Decoding parallelism benefits extraction.}
Semi-autoregressive decoding partitions the sequence into blocks of size $b$, decoding within each block in parallel while preserving order across blocks. Small $b$ approximates left-to-right autoregressive decoding (each block holds few tokens, committed in turn), whereas large $b$ recovers fully-parallel diffusion (the whole sequence is one bidirectional block). \Cref{fig:enron-block-size} shows recall increases with $b$, with the gains realized once $b$ leaves the AR-like regime. Restricting $b$ to small values recovers the more limited memorization profile, identifying bidirectional, fully-parallel denoising as a principal driver of infilling-based extraction.

\textbf{Sampling diversity slightly compromises extraction.}
Temperature $\tau$ controls how concentrated the per-position predictive distribution is: $\tau\!\to\!0$ approaches greedy decoding, while a higher $\tau$ flattens the distribution and injects stochasticity at committed tokens. \Cref{fig:enron-temperature} shows recall is relatively stable for $\tau$ below $1$, with greedy decoding yielding the strongest extraction for three of the four infilling modes. The effect is smaller in magnitude than the $T$ and $b$ sweeps because candidate positions already correspond to sharply peaked posteriors, and modest stochasticity has limited room to dilute the top-1 prediction.

\textbf{Confident remasking amplifies extraction.}
At each denoising step, the remasking strategy decides which tentative predictions to commit and which to re-mask for further refinement. \Cref{fig:enron-remasking} compares low-confidence remasking against random remasking on LLaDA-8B-FT: confidence-based remasking, which preserves high-confidence predictions and re-corrupts uncertain ones, yields markedly stronger extraction, while random remasking discards the model's confidence signal and slows convergence toward memorized targets. High confidence at a position is itself a marker of training-set provenance~\cite{ChenZDSFKRL26}, so committing those positions early lets the remaining trajectory consolidate around them and amplifies memorized content over the course of denoising.

Based on these observations, an adversary targeting open-weight DLMs can tune decoding parameters to maximize extractability, while a model owner, as the defender, can select appropriate parameters by balancing throughput, quality, and data leakage resistance.

\begin{figure}[htbp!]
    \vspace{-10pt}
    \centering
    \begin{minipage}[t]{0.49\textwidth}
        \centering
        \begin{subfigure}[t]{0.49\linewidth}
            \includegraphics[width=\linewidth]{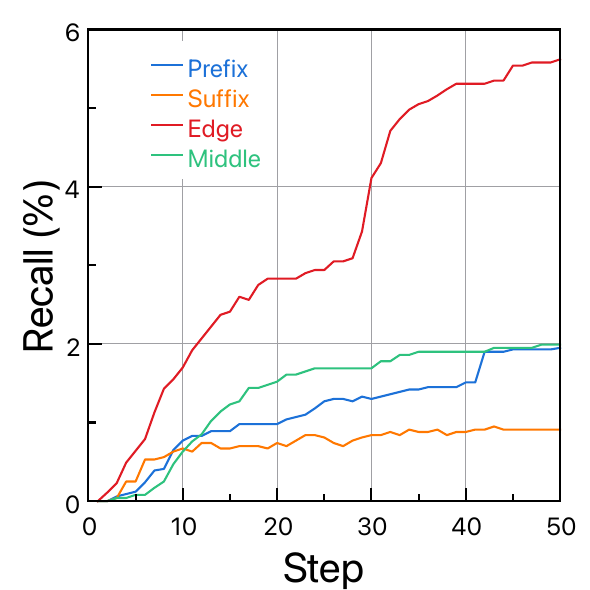}
            \caption{Email}
            \label{fig:llada-email-perstep}
        \end{subfigure}
        \hfill
        \begin{subfigure}[t]{0.49\linewidth}
            \includegraphics[width=\linewidth]{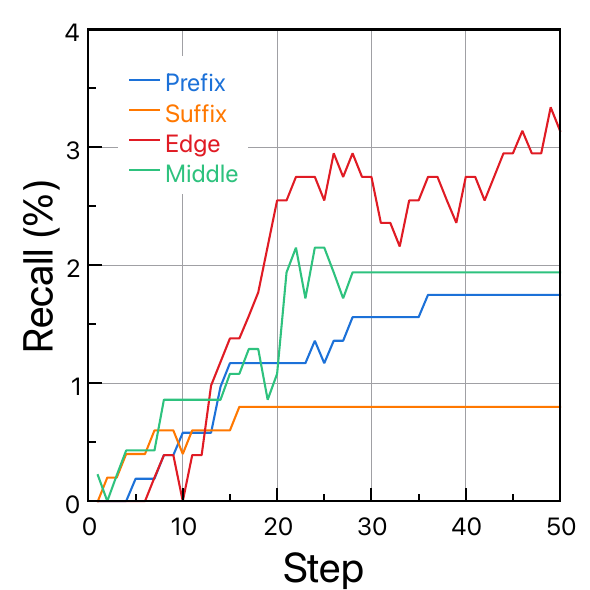}
            \caption{Phone}
            \label{fig:llada-phone-perstep}
        \end{subfigure}
        \caption{PII extraction trajectory during the decoding procedure.}
        \label{fig:pii-extraction-trajectory}
    \end{minipage}
    \begin{minipage}[t]{0.49\textwidth}
        \centering
        \begin{subfigure}[t]{0.49\linewidth}
            \includegraphics[width=\linewidth]{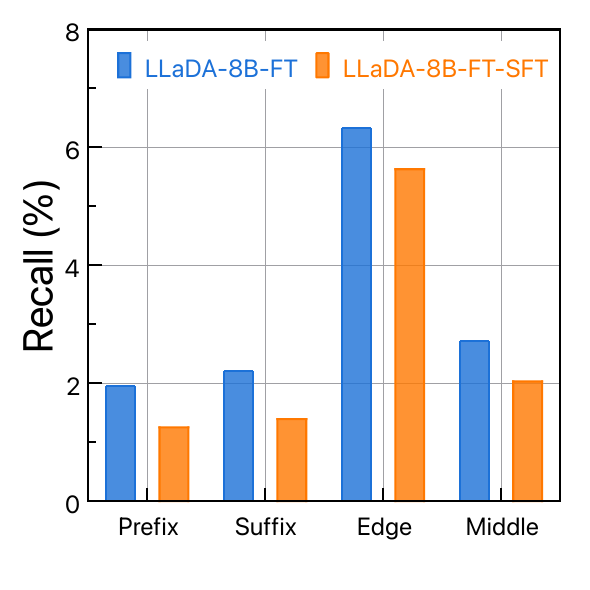}
            \caption{Email}
            \label{fig:llada-email-ft-tulu-sft}
        \end{subfigure}
        \hfill
        \begin{subfigure}[t]{0.49\linewidth}
            \includegraphics[width=\linewidth]{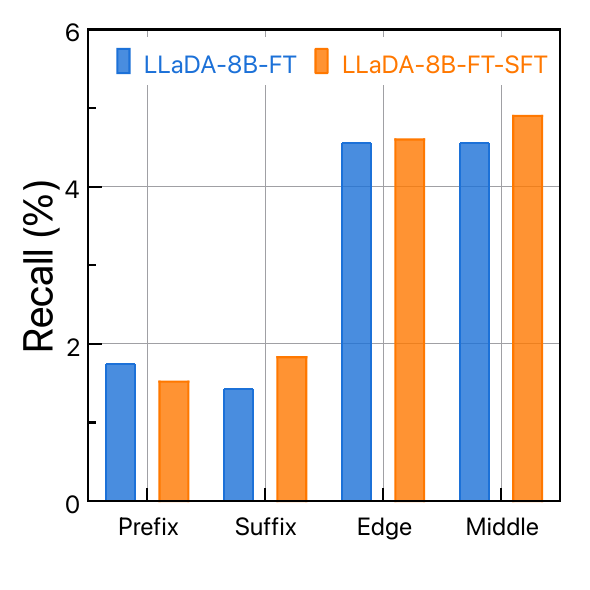}
            \caption{Phone}
            \label{fig:llada-phone-ft-tulu-sft}
        \end{subfigure}
        \caption{PII extraction results before and after SFT.}
        \label{fig:llada-enron-ft-tulu-sft}
    \end{minipage}
    \vspace{-10pt}
\end{figure}
\subsection{Extraction Trajectory}
\label{subsec:extraction-trajectory}
\Cref{fig:pii-extraction-trajectory} traces per-step PII recall on LLaDA-8B-FT. Across all four modes, the recall rises steadily through roughly the first 30 steps of the trajectory and then enters a regime of plateau or low-amplitude oscillation, with little further sustained gain in the remaining steps. A natural mechanism behind this saturation is that each step fills masked positions with the model's own predictions, so the decoding trajectory accumulates errors and progressively drifts from the ground-truth surroundings; once this drift is large enough, subsequent commitments are pulled toward generic completions rather than ground-truth PII, driving the marginal recall increment toward zero.

\subsection{Effect of SFT on PII Extractability}
\label{subsec:sft}
A typical deployment pipeline applies instruction tuning on top of domain fine-tuning. To test whether the second stage attenuates the memorization induced by the first, we continue to train LLaDA-8B-FT (Enron) with SFT on Tulu3 and re-run the infilling extraction on the original Enron set (\Cref{fig:llada-enron-ft-tulu-sft}).

PII extractability largely persists through SFT, and the relative ordering of modes is preserved across stages. SFT therefore redistributes the memorization budget without closing the structural leakage channel that infilling exposes. The attenuation is entity-type dependent. Email recall drops under every mode; phone recall, by contrast, is essentially unchanged or slightly higher in three of four modes. A hypothesis for this distinction is: emails lie within the natural-language distribution that SFT updates and are partially overwritten, while short idiosyncratic numeric strings are orthogonal to the instruction-following objective and survive intact.

\section{Conclusion}\label{sec:conclusion}

We introduced \emph{infilling extraction}, a mask-parameterized data-extraction protocol that subsumes \textit{prefix} conditioning and matches the bidirectional inductive bias of DLMs. On LLaDA-8B and Dream-7B, it surfaces extraction channels and poses data leakage risks invisible to \textit{prefix}-conditioned probing.

\textbf{Limitations and future work.} Our study has three main scope limitations. First, because the pre-training corpora of LLaDA and Dream are not publicly available, we induce memorization through fine-tuning on three representative datasets rather than measuring it directly on the released base checkpoints. Second, we examine only three exposure regimes and do not consider preference-based post-training, such as RLHF~\cite{ZhaoGZG25}. Third, since a strict comparison between large DLMs and ARMs is infeasible due to differences in architecture and training paradigm, we conduct our comparison under a scale-matched and tuning-matched setting.
Looking forward, the DLM area is evolving rapidly, with new training paradigms and decoding acceleration algorithms appearing at a fast pace~\cite{IsraeBG25,DeschTG25,LiC25,LiZMYYSLVL25}. Characterizing how emerging techniques interact with memorization is a natural next step. Beyond memorization itself, infilling extraction is a general primitive for probing what a bidirectional model has internalized. We see promising connections to adjacent safety problems, including MIAs~\cite{ChenZDSFKRL26}, unlearning, and uncertainty calibration~\cite{ChangYWCYTG26,QianTLYP26}.

\section*{Acknowledgements}
This work is supported in part by the Wallenberg Visiting Professor Program, the Natural Sciences and Engineering Research Council of Canada (grant number RGPIN-2026-04826), and the Government of Ontario (RE011-038).

\newpage
\bibliographystyle{plainnat}
\bibliography{reference.bib}

\newpage
\appendix

\section{Experimental Settings}\label{sec:appendix}

\subsection{Datasets}

Following prior work~\citep{ChenZDSFKRL26}, we fine-tune models on the \textit{full pile} subset of the MIMIR dataset, which contains 10K sequences. To evaluate verbatim extraction, we randomly sample 5,000 snippets of 100 tokens from the training set.
Across all models, snippets share the same starting position and span 100 tokens under each model's tokenizer, though their character lengths may vary slightly.

For PII extraction, we fine-tune models on 10,000 samples from the Enron dataset and sample 100-token snippets that contain PII entities in their masked portion. For instance, in the prefix-conditioned setting, snippets contain no PII in the first 50 tokens but include at least one PII entity in the last 50 tokens. All models share the same starting position for each sampled snippet. For each conditional mode, we construct 2,000 snippets for email address extraction and 450 snippets for phone number extraction. For \textit{targeted} infilling, we randomly mix snippets from the other four modes and construct sets of the same size.

For instruction tuning, we train on 10,000 samples from the Tulu3 dataset and sample 1,000 records in which both the prompt and response are under 100 tokens.

\subsection{Model Training}\label{app-subsec:models}

Following~\cite{ChenZDSFKRL26,ZhouCTS26}, we train each model for 5 epochs using the AdamW optimizer and a global batch size of 32. On Enron and MIMIR Pile, we apply full fine-tuning with a constant learning rate of $5\times 10^{-5}$and a weight decay of 0.1. For LoRA, we update an adapter of rank 64 using a learning rate of $2\times 10^{-4}$ under a cosine schedule with $5\%$ warmup, and a weight decay of 0.01. On Tulu3, we perform instruction tuning with a peak learning rate of $2\times 10^{-5}$ under a cosine schedule with $10\%$ warmup.

\subsection{Computation Platform}\label{app-subsec:compuation-platform}
All models are trained on a node containing 4 NVIDIA H100 80GB GPUs. Evaluation of the infilling extraction is conducted on a separate node equipped with 4 NVIDIA L40S GPUs.

\section{Additional Results}

\begin{figure}[htbp!]
    \centering
    \begin{subfigure}[t]{0.24\linewidth}
        \includegraphics[width=\linewidth]{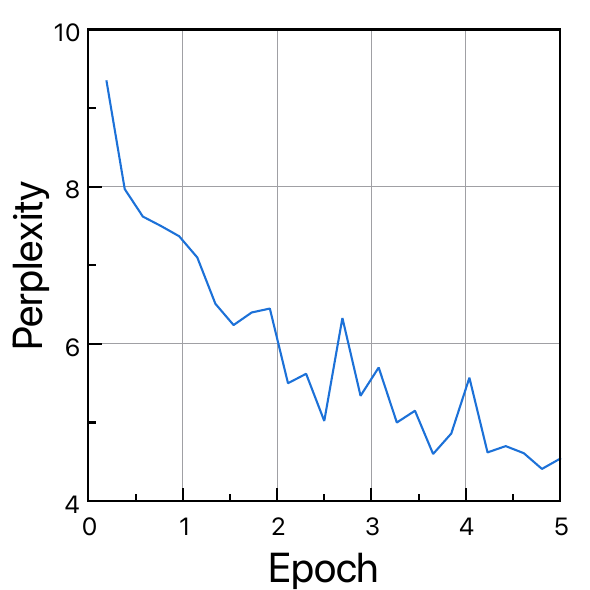}
        \caption{Training perplexity }
    \end{subfigure}
    \begin{subfigure}[t]{0.24\linewidth}
        \includegraphics[width=\linewidth]{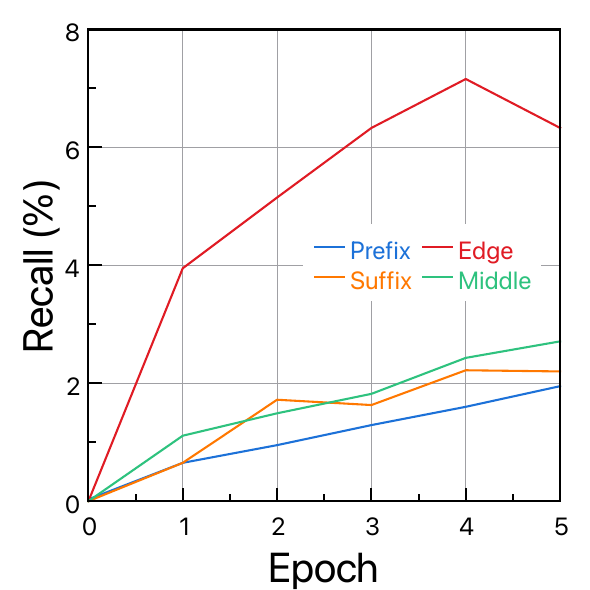}
        \caption{Recall of email}
    \end{subfigure}
    \begin{subfigure}[t]{0.24\linewidth}
        \includegraphics[width=\linewidth]{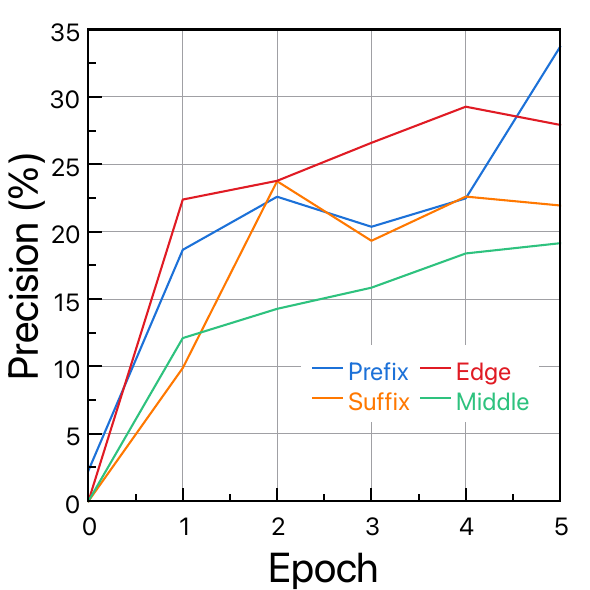}
        \caption{Precision of email}
    \end{subfigure}
    \caption{Extractability along the LLaDA-8B fine-tuning process on the Enron dataset.}
    \label{fig:llada-training}
\end{figure}
\subsection{Induced Memorization during Fine-tuning}
\Cref{fig:llada-training} shows the joint evolution of training perplexity, recall, and precision over the course of fine-tuning LLaDA-8B on the Enron Email dataset.
The perplexity curve descends from around 9.5 to 4.5 during the five-epoch fine-tuning. 
Before the fine-tuning, both the recall and precision of four extraction modes are near zero. For recall, \Cref{fig:llada-training} shows that the \textit{edge} increases the most among the four conditional modes and ends at above $6\%$ while other modes also reach around $2\%$. For precision, it shows that the four modes achieve final metrics between $18-35\%$ after fine-tuning.
\Cref{tab:llada-before-after-ft} compares the recall before and after fune-tuning on Enron.

\begin{table}[th]
\centering
\caption{PII recall (\%) of LLaDA before and after finetuning on Enron.}
\label{tab:llada-before-after-ft}
\begin{tabular}{llccccc}
\toprule
\textbf{PII Type} & \textbf{Model} & \textbf{Prefix} & \textbf{Suffix} & \textbf{Edge} & \textbf{Middle} & \textbf{Targeted} \\
\midrule
\multirow{2}{*}{Email Address}
  & LLaDA-Base & 0.03 & 0.00 & 0.00 & 0.00 & \textbf{1.37} \\
  & LLaDA-FT   & 1.95 & 2.20 & 6.33 & 2.71 & \textbf{40.50} \\
\midrule
\multirow{2}{*}{Phone Number}
  & LLaDA-Base & 0.00 & 0.00 & 0.00 & 0.00 & \textbf{0.30} \\
  & LLaDA-FT   & 1.75 & 1.43 & 4.56 & 4.56 & \textbf{28.90} \\
\bottomrule
\end{tabular}
\end{table}



\subsection{More Results of Different $k$}
We extend the context length $k$ from $50$ to the range [30, 70]. \Cref{tab:mask-k} shows the verbatim extraction ASR (\%) on LLaDA8B-FT and MIMIR. The edge mode remains the strongest of the five protocols.

\begin{table}[th]
\centering
\caption{Results of different masking strategies under varying $k$.}
\label{tab:mask-k}
\begin{tabular}{cccccc}
\toprule
$k$ & \textbf{Prefix} & \textbf{Suffix} & \textbf{Edge} & \textbf{Middle} & \textbf{Random} \\
\midrule
30 & 0.11 & 0.06 & \textbf{0.14} & 0.04 & \textbf{0.14} \\
40 & 0.16 & 0.13 & \textbf{0.28} & 0.09 & 0.19 \\
50 & 0.18 & 0.17 & \textbf{0.41} & 0.12 & 0.37 \\
60 & 0.27 & 0.27 & \textbf{0.65} & 0.15 & 0.58 \\
70 & 0.38 & 0.42 & \textbf{0.90} & 0.27 & 0.87 \\
\bottomrule
\end{tabular}
\end{table}

\subsection{Member vs.\ Non-member PII Extraction}
\label{app:nonmember}

To quantify how much of the extracted PII comes from noise or distributional guesses, we evaluate LLaDA-8B-FT on non-member Enron snippets that are never seen during fine-tuning, so any PII extracted from them reflects coincidental matches. As shown in Table~\ref{tab:nonmember}, recall on non-members is at least an order of magnitude lower than on members under every mode. The reported leakage therefore reflects genuine memorization rather than chance.

\begin{table}[h]
\centering
\caption{PII recall (\%) of LLaDA-8B-FT on member and non-member data of Enron.}
\label{tab:nonmember}
\begin{tabular}{llcccc}
\toprule
PII Type & Data & Prefix & Suffix & Edge & Middle \\
\midrule
\multirow{2}{*}{Email Address} & Non-member & 0.12 & 0.00 & 0.11 & 0.05 \\
                               & Member     & 1.95 & 2.20 & 6.33 & 2.71 \\
\midrule
\multirow{2}{*}{Phone Number}  & Non-member & 0.05 & 0.09 & 0.05 & 0.16 \\
                               & Member     & 1.75 & 1.43 & 4.56 & 4.56 \\
\bottomrule
\end{tabular}
\end{table}

\subsection{More Results of Decoding Parameters}
\begin{figure}[htbp!]
        \centering
        \begin{subfigure}[t]{0.24\linewidth}
            \includegraphics[width=\linewidth]{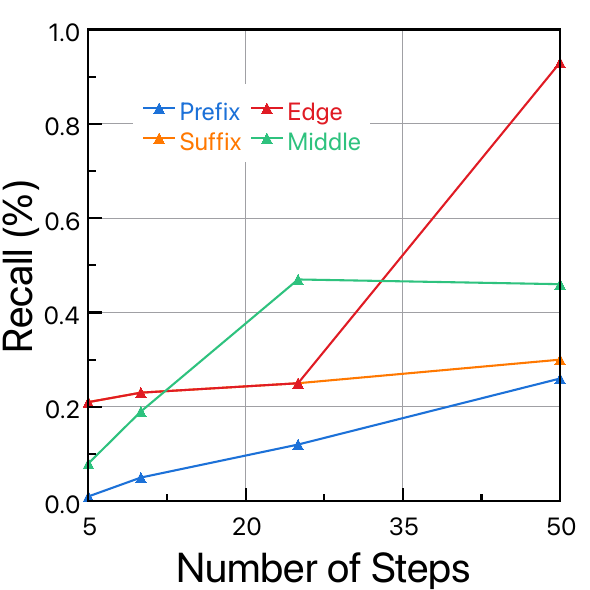}
            \caption{Number of steps $T$}
            \label{fig:dream-enron-step}
        \end{subfigure}
        \begin{subfigure}[t]{0.24\linewidth}
            \includegraphics[width=\linewidth]{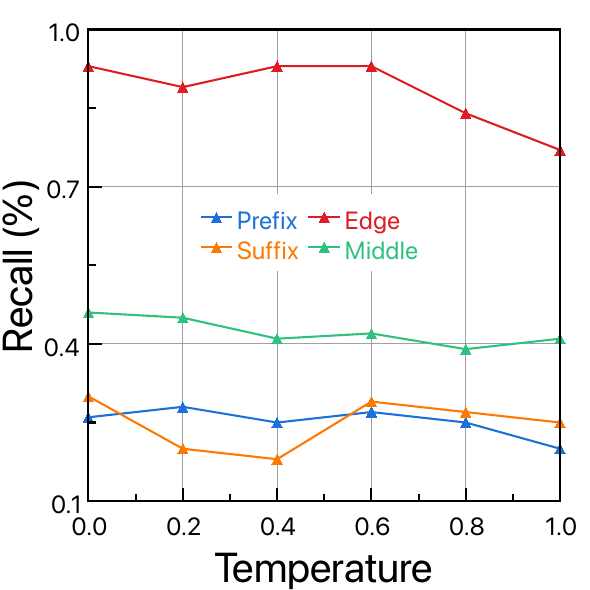}
            \caption{Temperature $\tau$}
        \label{fig:dream-enron-temperature}
        \end{subfigure}
        \begin{subfigure}[t]{0.24\linewidth}
            \includegraphics[width=\linewidth]{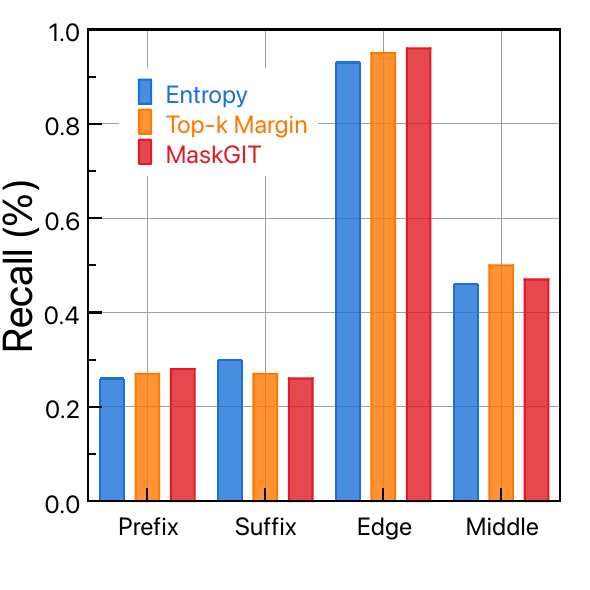}
            \caption{Remasking strategy}
        \label{fig:dream-enron-remasking}
        \end{subfigure}
        \caption{Influence of decoding parameters for Dream-7B-FT model on extracting email addresses from Enron dataset.}
        \label{fig:dream-influence-decoding-factors}
\end{figure}

\begin{figure}[htbp!]
        \centering
        \begin{subfigure}[t]{0.24\linewidth}
            \includegraphics[width=\linewidth]{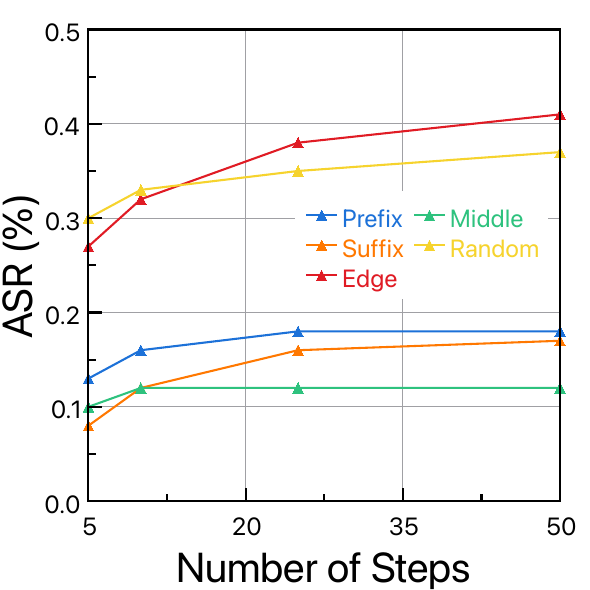}
            \caption{Number of steps $T$}
            \label{fig:mimir-step}
        \end{subfigure}
        \hfill
        \begin{subfigure}[t]{0.24\linewidth}
            \includegraphics[width=\linewidth]{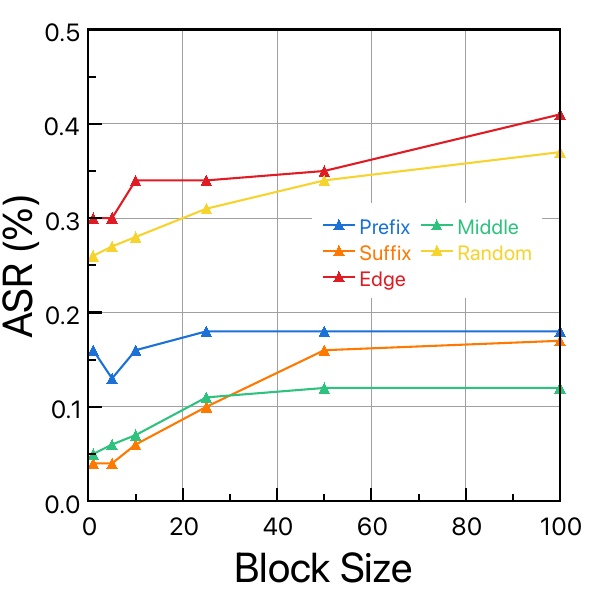}
            \caption{Block size $b$}
        \label{fig:mimir-block-size}
        \end{subfigure}
        \hfill
        \begin{subfigure}[t]{0.24\linewidth}
            \includegraphics[width=\linewidth]{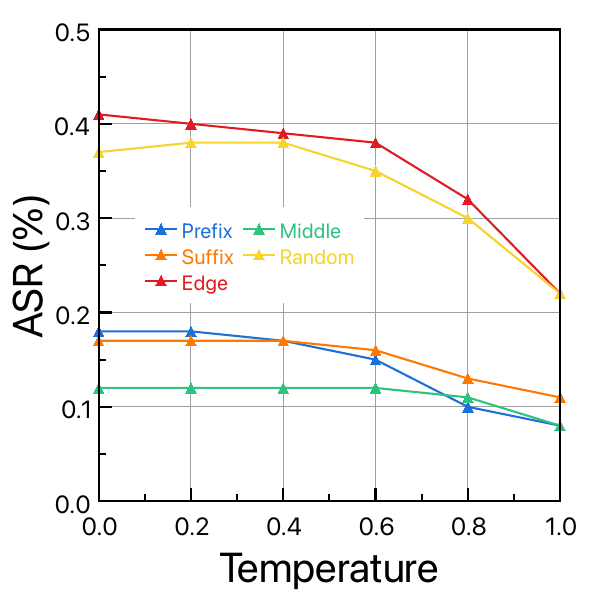}
            \caption{Temperature $\tau$}
        \label{fig:mimir-temperature}
        \end{subfigure}
        \hfill
        \begin{subfigure}[t]{0.24\linewidth}
            \includegraphics[width=\linewidth]{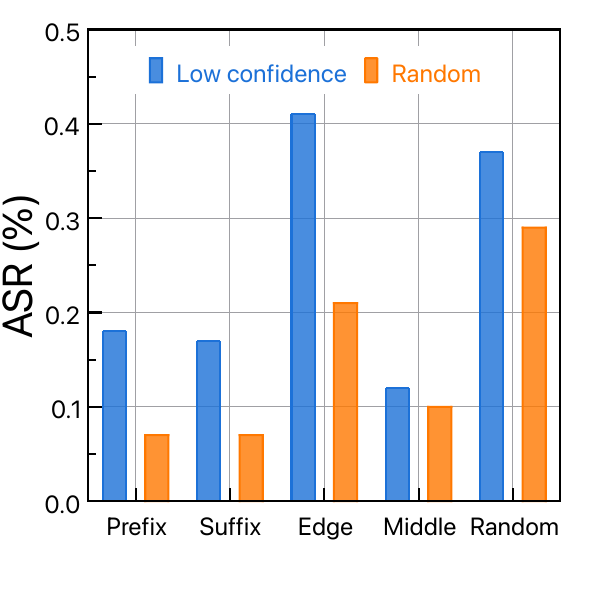}
            \caption{Remasking strategy}
        \label{fig:mimir-remasking}
        \end{subfigure}
        \caption{Influence of decoding parameters for LLaDA-8B-FT model on verbatim extraction on MIMIR Pile dataset.}
        \label{fig:mimir-influence-decoding-factors}
\end{figure}

Besides the discussion in \Cref{fig:influence-decoding-factors}, we provide additional results of Dream on Enron (\Cref{fig:dream-influence-decoding-factors}) and those of LLaDA on MIMIR Pile (\Cref{fig:mimir-influence-decoding-factors}).

As Dream does not support semi-autoregressive decoding, we investigate three decoding parameters on it: the number of decoding steps $T$, sampling temperature $\tau$, and remasking strategy.
We evaluate three strategies, including \emph{entropy-based}, \emph{top-$k$ margin}, and \emph{MaskGIT}.
\Cref{fig:dream-influence-decoding-factors} shows that higher $T$ increases the extraction rate while higher temperature slightly compromises the recall, which stays consistent with previously observed patterns for LLaDA.

\Cref{fig:mimir-influence-decoding-factors} further confirms on MIMIR Pile that higher $T$, higher $b$ and lower $\tau$ are preferred for higher extractability of LLaDA.

\section{More Related Work}
Carlini et al.~\cite{CarliTWJHLRBSEOR21} first showed that ARMs regurgitate verbatim training sequences, including PII, when prompted with prefixes. Later work quantified how memorization scales with model size, data duplication, and context length~\cite{carlini2023quantifying,LeeINZECC22}. It also improved extraction through better sampling and ranking strategies~\cite{YuPLDKHLY23}, replaced the deterministic verbatim criterion with probabilistic extraction~\cite{HayesSCYSNCLC25}, and extended extraction to aligned production models~\cite{NasrCHJCICWTL23} and fine-tuned models~\cite{ZengLRLXHXWTY24}. Almost all of this work follows the prefix-conditioned protocol, which matches left-to-right generation; Xiong et al.~\cite{XiongZPS25} survey the area.

\section{Broader Impacts}\label{app-sec:broader-impacts}
DLMs offer higher generation efficiency while achieving generation quality comparable to ARMs, making them promising foundation models for future products. 
In this paper, we propose a data extraction framework tailored to DLMs and evaluate the data-leakage risks induced by infilling extraction. 
We show that, in certain scenarios, the data-leakage risks posed by DLMs have been underestimated. In particular, when companies or government agencies train models on sensitive data but subsequently release (e.g., for transparency or regulatory compliance) redacted versions of the training data, targeted infilling can effectively recover the sensitive data memorized by the models. Our findings shed light on a previously unexplored risk surface of DLMs.
As a potential mitigation, we further demonstrate the influence of decoding parameters, which model owners can tune to balance throughput, generation quality, and data-leakage resistance.



\end{document}